\documentclass{article}
\usepackage{spconf,amsmath,graphicx}
\usepackage{url}
\usepackage{enumitem}
\setlist{nosep, leftmargin=14pt}
\usepackage{tcolorbox}

\usepackage{mwe} 

\title{Generating customized prompts for Zero-Shot Rare Event Medical Image
Classification using LLM}
\name{Payal Kamboj, Ayan Banerjee, Bin Xu and Sandeep Gupta}

\address{School of Computing and Augmented Intelligence, \\Arizona State University, Tempe, USA}
\begin{document}
%
\maketitle
\begin{abstract}
Rare events, due to their infrequent occurrences, do not have much data, and hence deep learning techniques fail in estimating the distribution for such data. Open-vocabulary models represent an innovative approach to image classification. Unlike traditional models, these models classify images into any set of categories specified with natural language prompts during inference. These prompts usually comprise manually crafted templates (e.g., `a photo of a \{\}') that are filled in with the names of each category. This paper introduces a simple yet effective method for generating highly accurate and contextually descriptive prompts containing discriminative characteristics. Rare event detection, especially in medicine, is more challenging due to low inter-class and high intra-class variability. To address these, we propose a novel approach that uses domain-specific expert knowledge on rare events to generate customized and contextually relevant prompts, which are then used by large language models for image classification. Our zero-shot, privacy-preserving method enhances rare event classification without additional training, outperforming state-of-the-art techniques. Code available at \texttt{\url{https://github.com/payalkamboj/CuPKL}.}
\end{abstract}

\section{Introduction}
\label{section1}
\vspace{-1.5mm}
~\textit{``Rare events are extremely infrequent events whose characteristics make them or their consequences highly valuable.
Such events appear with extreme scarcity and are hard to
predict, although they are expected eventually''}~\cite{sokolova2010evaluation}. Despite the rarity, these instances are highly significant as they contain crucial information. Detecting such rare events requires advanced methods that account for their infrequency, unique features, and positive-negative evidences, since the true rare event distribution cannot be fully captured from these sparse observations using purely statistical methods~\cite{abubakar2024systematic}. In the current era of large language and vision models (LLM/LVM), which are extensively trained on public data and demonstrate impressive reasoning capabilities~\cite{wei2022chain}, both pattern recognition and deductive reasoning (i.e. their capacity to
follow instructions in reasoning tasks)~\cite{cheng2024inductivedeductiverethinkingfundamental}, we explore whether these models can be adapted to effectively classify rare events despite the limited availability of event-specific data.

Open-vocabulary models demonstrate high image classification accuracy across various datasets, even without labeled training data specific to those tasks~\cite{pratt2023does}. They achieve this by leveraging extensive collections of image-text pairs from the internet, learning to associate images with their corresponding captions. This method offers significant flexibility during inference. Unlike traditional models, these models classify images by assessing the similarity between an image and a caption. During inference, a caption or `prompt' is generated for each desired category, enabling the model to match each image with the most suitable prompt (standard zero-shot\footnote{Zero-shot image classification is the process of categorizing images into classes that the model hasn’t seen during training}). The ability of open-vocabulary models to accurately match text and images stems from the diverse visual and linguistic patterns they have encountered during training, allowing them to generalize well to unseen categories. As a result, they can draw upon these learned associations to align images and text efficiently, even when faced with unfamiliar data. This allows for on-the-fly category selection and adjustments without the need for additional training. However, this innovative approach introduces a new challenge:

\textit{How can we effectively describe an image category using natural language prompts for categories that the model has likely never seen before in text or visual?}

The conventional method involves manually crafting prompt templates (e.g.,``a photo of a \{\}”), generating a natural language label for each category in the dataset, and creating a collection of prompts for each category by populating each template with the corresponding labels. Subsequently, image embeddings are compared to the nearest set of prompt embeddings and assigned the category linked to that set of prompts~\cite{radford2021learning}. 
However, these methods have drawbacks: \\
\noindent a) The standard prompts, such as ``a photo of \{\}" of standard zero shot lack domain-specific information, which is crucial in medical imaging as public datasets are limited,
\begin{figure*}[t]
\centering
\includegraphics[width=0.9\textwidth,clip=true,trim= 0 0 0 20]{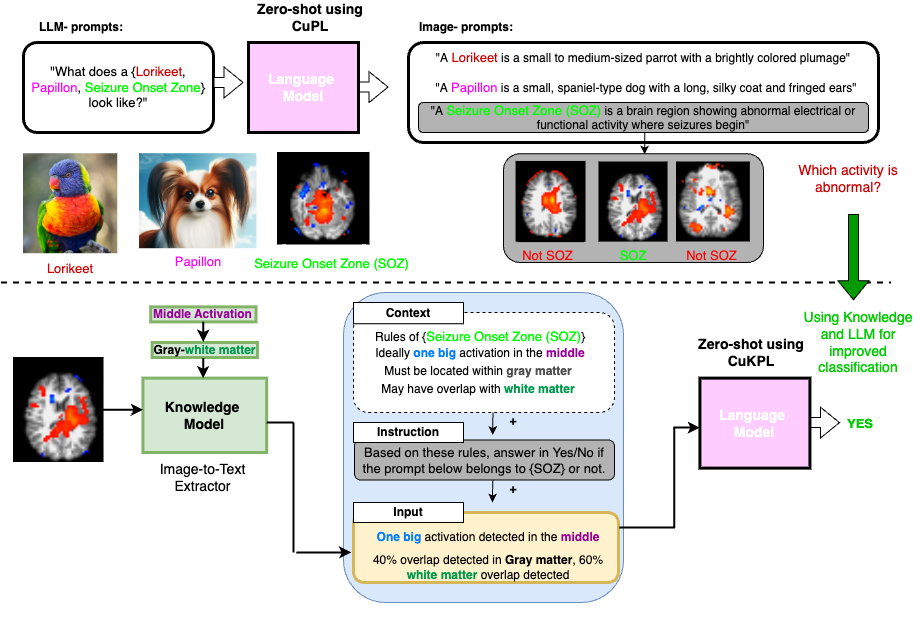}
\caption{\textbf{Example CuPL LLM-prompts, Image prompts (Top), and CuKPL LLM prompts and classification (Bottom).}}
\label{fig:1}
\end{figure*}
b) These methods are usually evaluated on datasets where categories have been indirectly encountered during training. For example, a model trained on images of dogs with captions like ``Aussie the pup" is likely to classify the category ``Dog" correctly during testing due to semantic similarities with the label~\cite{radford2021learning}. In contrast, for rare events, such as seizure onset zones (SOZ), these models struggle to generalize due to insufficient data and a lack of relevant associations in the training corpus, c) Standard prompt templates lack detailed descriptions which are crucial for fine-grained classification~\cite{pratt2023does}. \\

\vspace{-3.5mm}
To address these challenges, we propose Customized Prompts via Knowledge and Language Models (CuKPL), which combines domain-specific knowledge of rare events with image-specific information in an instruction-input format consistent with the tuning of most LLMs~\cite{zhang2023instruction,banerjee2024cps}. CuKPL generates \textbf{Human Knowledge-Embodied Textual Prompts (HKETP)}, descriptive of rare event categories, by extracting and structuring insights from technical literature, often expressed in vague natural language. This process yields two key outcomes: a) A structured format that encapsulates the relevant insights and findings derived from the literature (\textbf{Context}, Fig.~\ref{fig:1}), and b) Knowledge components to analyze and quantify different aspects of knowledge from the image to classify, leading to HKETP (\textbf{Input}, Fig.~\ref{fig:1}). The advantage of using CuKPL is that it enables effective classification using only the text modality, eliminating the need for an image encoder or LVM. The context prompt, and input prompt is integrated with an \textbf{instruction} prompt (Fig.~\ref{fig:1}) and then input into an LLM, which outputs the image category.

Rare event detection in biomedical imaging is challenging due to high intra-class and low inter-class variability. Discriminative knowledge rules play a critical role in distinguishing rare events from visually similar instances. Extensive experiments show that CuKPL surpasses state-of-the-art (SOTA) methods, operating within a zero-shot framework that preserves data privacy by ensuring the LLM never accesses the image, is computationally efficient by eliminating the need for training , and avoids reliance on synthetic data generation, which is ineffective in low-data scenarios~\cite{lane2010balancing}.
\begin{figure*}[t]
\centering
\includegraphics[width=0.92\textwidth,clip=true,trim= 0 0 0 45]{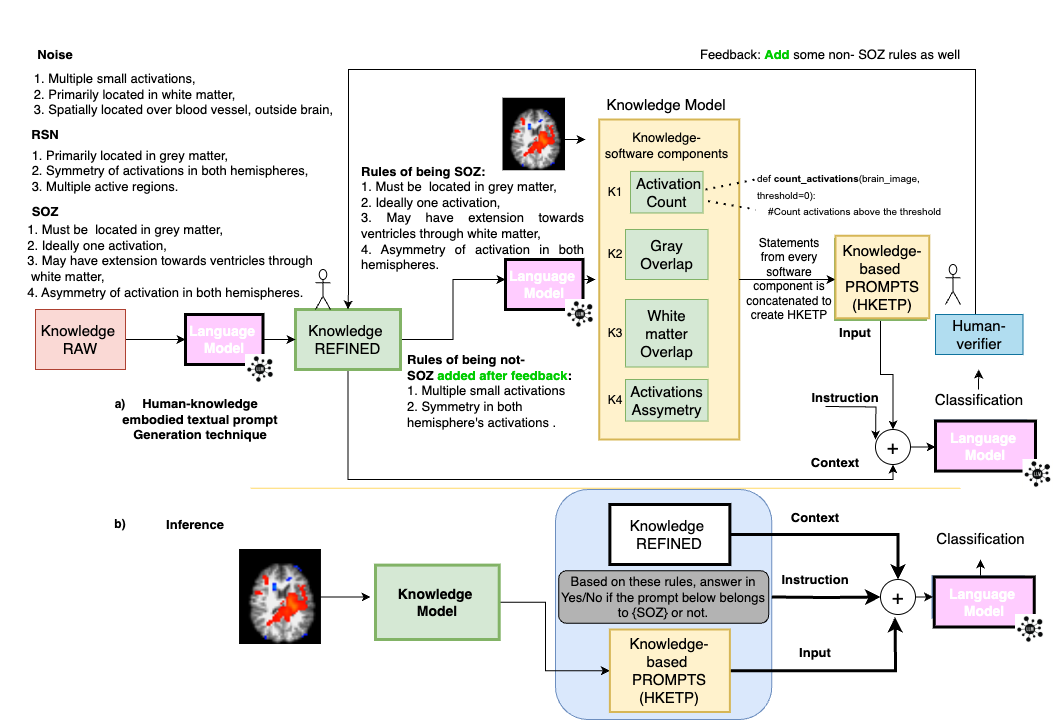}
\caption{\textbf{a) HKETP Generation:} Raw knowledge refined by LLM is encoded using image processing techniques in knowledge model, producing HKETP. Refined knowledge (Context), HKETP (Input) along with Instruction are used by LLM for classification, with feedback from a human verifier. \textbf{b) Inference:} Given an image, Knowledge model outputs input HKETP which combined with refined knowledge context and instruction is collectively used by LLM to classify rare event.}
\label{fig:2}
\end{figure*}
\vspace{-2mm}
\section{Related Work}
\label{RW}
\vspace{-1mm}
\textbf{Standard zero-shot:} Standard zero-shot setup uses predefined templates to infer a category without any training~\cite{radford2021learning,pham2023combinedscalingzeroshottransfer,jia2021scalingvisualvisionlanguagerepresentation,yu2022cocacontrastivecaptionersimagetext}. However, these models are trained on large, publicly available datasets, which often lack representation of medical data. \textbf{Customized prompts using LLM} such as CuPL~\cite{pratt2023does}, integrates open-vocabulary models with LLMs to generate customized prompts. 
However, in medical domain where data is scarce, LLMs struggle to produce detailed and accurate prompts, leading to suboptimal performance. \textbf{SOTA medical imaging classification} include ViTs, 2D-CNNs, which are highly dependent on large datasets. Alternatives have also emerged that leverage domain knowledge alone~\cite{banerjee2023automated,hossain2023edgcon} or in combination with DL~\cite{10518119,kamboj2023merging,banerjee2024framework,kamboj2024storm,gupta2024systems}.~Combining domain knowledge with AI has shown superior performance compared to standalone methods. However, they are heavily data-reliant.

\vspace{-2mm}
\section{Methods}
\vspace{-2mm}
Fig. \ref{fig:2} shows our overall methodology, which extends the SOTA customized prompt driven image recognition method by integrating expert knowledge into the prompts, structured in an instruction-input format that LLMs are trained to process. The overall process is divided in two parts: 
 \begin{table*}
 \scriptsize
\centering
\caption{Performance results of SOTA models and CuKPL on Center A (Top-half). Performance results for all models trained on data from Center A and evaluated on unseen data from Center B (Bottom-half). MM: Machine marked}\label{tab1}
\begin{tabular}{|l|l|l|l|l|l|l|}
\hline
\multicolumn{2}{|l|}{\textbf{Method}} &  \textbf{Accuracy} & \textbf{Precision} &\textbf{Sensitivity} &\textbf{F1-score} &\textbf{MM SOZs}\\
\hline
\multicolumn{2}{|l|}{DL-2D CNN}   &46.1\%  &88.8\% &48.9\% &63.0\% &10\\
\hline

\multicolumn{2}{|l|}{DL-ViT}   &34.6\%  &85.7\% &36.7\% &51.3\% &13\\

\hline
\multicolumn{2}{|l|}{Knowledge-based system (EPIK)}   &75.0\%  &92.8\% &79.5\% &85.6\% &43\\
\hline

\multicolumn{2}{|l|}{ Supervised-Knowledge-based system}   &50.0\%  &89.6\% &53.6\% &67.0\% &6\\
\hline

\multicolumn{2}{|l|}{Knowledge and DL (DeepXSOZ)}   &84.6\%  &93.6\% &89.7\% &91.6\% &18\\

\hline
\multicolumn{2}{|l|}{Standard zero shot (LVM-CLIP)}   &All ICs classified as Noise  &- &- &- &-\\
\hline
\multicolumn{2}{|l|}{CuPL}   & Most ICs classified as Noise, few as RSN  &- &- &- &-\\
\hline

\multicolumn{2}{|l|}{ \textbf{CuPKL GPT-4o}}   &\textbf{88.4}\%  &\textbf{93.8}\% &\textbf{93.8}\% &\textbf{93.8}\% &\textbf{28}\\
\hline
\hline

\multicolumn{2}{|l|}{DL-2D CNN Center B}   &67.7\%  &87.5\% &75.0\% &80.7\% &14\\

\hline
\multicolumn{2}{|l|}{DL-ViT Center B}   &12.9\%  &57.1\% &14.2\% &22.7\% &6\\
\hline
\multicolumn{2}{|l|}{Knowledge-based system (EPIK) Center B}   &52.1\%  &85.7\% &58.1\% &68.5\% &27\\
\hline
\multicolumn{2}{|l|}{Supervised-Knowledge-based system, Center B}   &83.8\%  &89.6\% &92.8\% &91.2\% &21\\
\hline

\multicolumn{2}{|l|}{Knowledge and DL (DeepXSOZ) Center B }   &90.3\%  &90.3\% &100\% &94.9\% &28\\
\hline
\multicolumn{2}{|l|}{CuPKL GPT-4o Center B }   &70.0\%  &90.3\% &75.0\% &82.3\% &14\\

\hline
\end{tabular}
\end{table*}
  
\noindent{\textbf{a) HKETP Generation}} As illustrated in Fig.~\ref{fig:2} (top), raw textual knowledge from literature is initially refined using an LLM to capture non-overlapping, discriminative knowledge specific to the rare class. This refined knowledge is then structured into knowledge components and encoded (using software functions), generating rule-based text outputs from the image. Each knowledge component is encoded with the help of LLM to produce targeted prompts, each with a degree of satisfiability based on the input image. These generated texts are integrated (Fig.~\ref{fig:2}), creating HKETP (input), which, along with the refined knowledge (context) and instruction, is then utilized by the LLM for final classification. Following classification, a human verifier provides feedback and updates, based on one patient's data, to the LLM such as add some more contextual knowledge, update thresholds in the knowledge components, to ensure precise knowledge encoding and improvement in the classification output. \\
\noindent{b) \textbf{Inference:}} Fig.~\ref{fig:2} (bottom) illustrates the LLM-based classification process, where the HKETP input (generated by the knowledge model is combined with the final refined knowledge context (obtained after iterations of human verifier's feedback) and instruction. These three are then fed to the LLM as a prompt to classify the presence of target rare class. This approach leverages the instruction-tuned capabilities of LLMs, allowing any LLM fine-tuned on the generic instruction-input-output format to be used for classification.
\vspace{-0.1 in}\subsection{Application Domain: Rare SOZ Detection}

\vspace{-2.5mm}

The raw 4D spatio-temporal fMRI data is processed using Independent component (IC) analysis to get three kinds of ICs: noise, which captures measurement artifacts, resting state networks (RSN) that capture normal brain activity, and SOZ ICs (rare class). fMRI yields ~150 ICs per patient with ($<$ 10\%) associated with SOZ~\cite{10518119}~\cite{banerjee2023automated}~\cite{kamboj2024expert}~\cite{boerwinkle2017correlating}. The problem is: given an IC we need to classify if it belongs to SOZ. 

\noindent \textbf{Solution:}  Few key knowledge instances on SOZ include: a) Single activation within the brain~\cite{10518119}, b) Activation in gray matter, may overlap with white matter~\cite{10518119},
c) Assymetry in hemisphere's activations~\cite{10518119} etc. This raw knowledge was compiled from the literature~\cite{10518119},~\cite{banerjee2023automated}, and further refined by LLM to focus on discriminative knowledge (Figure~\ref{fig:2}, Top). Each specific knowledge component (K1, K2,..., KN) was encoded using the LLM, triggering distinct prompts when the corresponding image aligned with domain knowledge. These knowledge-specific components, forming the HKETP (input), are integrated with the refined knowledge relevant to the category of interest (context). For inference, a spatial IC image is given to the knowledge model, generating HKETP. HKETP input along with instruction and refined knowledge context is used by LLM to classify the image as SOZ or not SOZ.
\begin{tcolorbox}[colback=gray!10, colframe=gray!80, title=Customized prompt example by CuKPL]
\scriptsize
\textbf{Rules of seizure onset zone (SOZ) are as follows:}
\begin{enumerate}
    \item Detection of 1 big red activation.
    \item Big red activation has asymmetry in hemisphere.
     \item Big red activation has NO similar shape.
    
\end{enumerate}

\textbf{ Rules of being not SOZ:}
\begin{enumerate}
    \item Lots of small/medium activations detected.
    \item Symmetry in activations in both hemispheres.
    \item Detection of 1 big red activation has majority overlap in gray matter.
\end{enumerate}

Based on rules, answer in \textbf{YES/NO} if the prompt below belongs to SOZ. If uncertain, respond \textbf{NO}.

1 big red activation(s) detected!
4 small red activation(s) detected!
Big red activation has Percentage overlap with gray matter: 92.5\%, with white matter: 0.0\%,
The big red activation has asymmetrical shapes.
\end{tcolorbox}

\section{Experimental Setup and Evaluation}

\vspace{-1mm}
Data collected from two centers, A and B, in compliance with IRB protocols. Center A, PCH, Phoenix, has 52 pediatric patients (23 Male, 29 Female, ages 3 months to 18 years) with 5,616 images (2,873 Noise, 2,427 RSN, 316 SOZ), acquired with 3T Philips Ingenuity. Center B, UNC, Chapel hills, has 31 patients (14 Male, 17 Female, ages 2 months to 62 years) with 2,364 images (1,090 Noise, 1,072 RSN, 202 SOZ), acquired using a Siemens MAGNETOM Prisma FIT. 

\noindent{\textbf{Evaluation}}: We implemented CuKPL using GPT-4o and compared it with ViT, 2D-CNN, and knowledge-based methods (EPIK~\cite{banerjee2023automated}, Hunyadi et al.~\cite{hunyadi}) to assess its effectiveness in rare class classification.  We also evaluated the integration of knowledge with DL\cite{10518119} and SOTA zero-shot image classification models using open-vocabulary systems. Evaluations included: (1) Training on center A and evaluating on center A with Leave-One-Out Cross-Validation, and (2) Training on center A and testing on unseen center B. Please note that CUKPL doesn't require any training. It just needs one patient's data during HKETP generation process to tune knowledge model. We also assess the potential effort reduction for the surgical team by comparing the ICs that require review by the surgical expert to those that would need manual sorting using machine marked SOZs~\cite{10518119}.\\
\noindent{\textbf{Results Analysis:}} SOTA DL techniques like 2D-CNNs and ViT, yielded suboptimal results (Table~\ref{tab1}). Knowledge-based approaches, such as EPIK~\cite{banerjee2023automated} and supervised knowledge-based systems~\cite{hunyadi}, offered improvements suggesting that human expert knowledge is valuable than pure data driven learning. Combining knowledge with DL achieved optimal results but remained data-dependent, non-zero-shot, and required extensive training. To address these limitations, we evaluated standard zero-shot methods like CLIP and CUPL, but neither possessed knowledge of rare medical events. CuKPL, on the other hand, which achieved a 3.8\% accuracy and 2.2\% F1 score improvement on Center A data over SOTA DeepXSOZ~\cite{10518119}. In single-domain generalization, CuKPL showed strong performance on Center B, with an F1 score of 82.3\% and 70\% accuracy. The integration of knowledge and DL proved most effective across evaluations. To further enhance CuKPL's performance on datasets from other centers, it would still require data from one patient at that center to adjust the thresholds in the knowledge model. As such, it would be more effective for domain adaptation than for domain generalization.  The MM SOZs of 28 show a significant reduction in manual evaluation by neurosurgeons, making them ideal for clinical use.
 
\vspace{-2mm}
\section{Conclusion}
\vspace{-2mm}
Traditional DL methods struggle with limited data, while purely knowledge-based approaches fall short due to vague or incomplete domain knowledge. Our framework, CuKPL, addresses these challenges by generating customized, knowledge-rich prompts that, in combination with contextual knowledge and instructions, enable effective image classification using LLMs. Extensive testing demonstrates that CuKPL outperforms SOTA methods and achieves single-domain generalization across datasets. However, it shows greater potential as a domain adaptation for datasets from different centers, which is a future work. Additionally, this solution generalizes well to other medical image classification tasks like proliferative diabetic retinopathy grading as well.
\\

\noindent\textbf{Ethical statement} This work adheres to ethical standards by prioritizing the responsible use of AI in medical applications, ensuring patient data privacy. The integration of human expertise with machine learning aims to enhance diagnostic accuracy while remaining mindful of AI's limitations in sensitive, data-limited contexts.\\
\noindent\textbf{Conflict of Interest}: The authors declare no conflicts of interest.
\bibliographystyle{IEEEbib}
\bibliography{strings,arxiv}

\end{document}